\documentclass[a4paper,11pt]{article}
\usepackage{acl-ijcnlp2009}
\usepackage{times} 
\usepackage{url} 
\usepackage{latexsym} 
\usepackage{multicol} 
\usepackage{mdwlist}
\usepackage{covington}

\title{An Annotation Scheme for Reichenbach's Verbal Tense Structure}

\author{Leon Derczynski \and Robert Gaizauskas\\
  Department of Computer Science\\
  University of Sheffield, UK\\
  \tt{\{leon,robertg\}@dcs.shef.ac.uk}
}

\begin{document}

\maketitle

\begin{abstract}
In this paper we present RTMML, a markup language for the tenses of verbs and temporal relations between verbs. There is a richness to tense in language that is not fully captured by existing temporal annotation schemata. Following Reichenbach we present an analysis of tense in terms of abstract time points, with the aim of supporting automated processing of tense and temporal relations in language. This allows for precise reasoning about tense in documents, and the deduction of temporal relations between the times and verbal events in a discourse. We define the syntax of RTMML, and demonstrate the markup in a range of situations.
\end{abstract}

\section{Introduction}

In his 1947 account, Reichenbach offered an analysis of the tenses of verbs, in terms of abstract time points. Reichenbach details nine tenses (see Table~\ref{tab:tenses}). The tenses detailed by Reichenbach are past, present or future, and may take a simple, anterior or posterior form.  In English, these apply to single verbs and to verbal groups (e.g. \emph{will have run}, where the main verb is \emph{run}). 

To describe a tense, Reichenbach introduces three abstract time points. Firstly, there is the speech time, $S$. This represents the point at which the verb is uttered or written. Secondly, event time $E$ is the time that the event introduced by the verb occurs. Thirdly, there is reference time $R$; this is an abstract point, from which events are viewed. In Example~\ref{exp:simple}, speech time $S$ is when the author created the discourse (or perhaps when the reader interpreted it). Reference time $R$ is \emph{then} -- an abstract point, before speech time, but after the event time $E$, which is the leaving of the building. In this sentence, one views events from a point in time later than they occurred.

\begin{example}
\emph{By then, she had left the building.}
\label{exp:simple}
\end{example}

While we have rich annotation languages for time in discourse, such as TimeML\footnote{http://www.timeml.org; \newcite{boguraev2005timeml}.} and TCNL\footnote{See~\newcite{han2006language}.}, none can mark the time points in this model, or the relations between them. Though some may provide a means for identifying speech and event times in specific situations, there is nothing similar for reference times. All three points from Reichenbach's model are sometimes necessary to calculate the information used in these rich annotation languages; for example, they can help determine the nature of a temporal relation, or a calendrical reference for a time. We will illustrate this with two brief examples.

\begin{example}
\emph{By April 26$^{\textrm{th}}$, it was all over.}
\label{exp:normalisation}
\end{example}

In Example~\ref{exp:normalisation}, there is an anaphoric temporal expression describing a date. The expression is ambiguous because we cannot position it absolutely without an agreed calendar and a particular year. This type of temporal expression is interpreted not with respect to speech time, but with respect to reference time~\cite{ahn2005towards}. Without a time frame for the sentence (presumably provided earlier in the discourse), we cannot determine which year the date is in. If we are able to set bounds for $R$ in this case, the time in Example~\ref{exp:normalisation} will be the April $26^{th}$ adjacent to or contained in $R$; as the word \emph{by} is used, we know that the time is the April $26^{th}$ following $R$, and can normalise the temporal expression, associating it with a time on an absolute scale.

Temporal link labelling is the classification of relations between events or times. We might say an event of \emph{the airport closed} occurred \textbf{after} another event of \emph{the aeroplane landed}; in this case, we have specified the type of temporal relation between two events. This task is difficult to automate~\cite{verhagen2010semeval}. There are clues in discourse that human readers use to temporally relate events or times. One of these clues is tense. For example:

\begin{example}
\emph{John told me the news, but I had already sent the letter.}
\label{exp:permanence}
\end{example}

Example~\ref{exp:permanence} shows a sentence with two verb events -- \emph{told} and \emph{had sent}. Using Reichenbach's model, these share their speech time $S$ (the time of the sentence's creation) and reference time $R$, but have different event times. In the first verb, reference and event time have the same position. In the second, viewed from when John told the news, the letter sending had already happened -- that is, event time is before reference time. As reference time $R$ is the same throughout the sentence, we know that the letter was sent before John mentioned the news. Describing $S$, $E$ and $R$ for verbs in a discourse and linking these points with each other (and with times) is the only way to ensure correct normalisation of all anaphoric and deictic temporal expressions, as well as enabling high-accuracy labelling of some temporal links.

Some existing temporal expression normalisation systems heuristically approximate reference time. GUTime~\cite{mani2000robust} interprets the reference point as ``the time currently being talked about", defaulting to document creation date. Over 10\% of errors in this system were directly attributed to having an incorrect reference time, and correctly tracking reference time is the only way to resolve them. TEA~\cite{han2006language} approximates reference time with the most recent time temporally before the expression being evaluated, excluding noun-modifying temporal expressions; this heuristic yields improved performance in TEA when enabled, showing that modelling reference time helps normalisation. HeidelTime~\cite{stroetgen2010heideltime} uses a similar approach to TEA but does not exclude noun-modifying expressions.

The recently created WikiWars corpus of TIMEX2 annotated text prompted the comment that there is a ``need to develop sophisticated methods for temporal focus tracking if we are to extend current time-stamping technologies"~\cite{mazur2010wikiwars}. Resources that explicitly annotate reference time will be direct contributions to the completion of this task.

\newcite{elson2010tense} describe how to relate events based on a ``perspective" which is calculated from the reference and event times of an event pair. They construct a natural language generation system that requires accurate reference times in order to correctly write stories. \newcite{portet2009automatic} also found reference point management critical to medical summary generation.

These observations suggest that the ability to automatically determine reference time for verbal expressions is useful for a number of computational language processing tasks. Our work in this area -- in which we propose an annotation scheme including reference time -- is a first step in this direction.

In Section~\ref{model} we describe some crucial points of Reichenbach's model and the requirements of an annotation schema for tense in natural language. We also show how to reason about speech, event and reference times. Then, in Section~\ref{rtmml}, we present an overview of our markup. In Section~\ref{examples} we give examples of annotated text (fictional prose and newswire text that we already have another temporal annotation for), event ordering and temporal expression normalisation. Finally we conclude in Section~\ref{conclusion} and discuss future work.

\section{Exploring Reichenbach's model}
\label{model}

Each tensed verb can be described with three points; speech time, event time and reference time. We refer to these as $S$, $E$ and $R$ respectively. Speech time is when the verb is uttered. Event time is when the action described by the verb occurs. Reference time is a viewpoint from where the event is perceived. A summary of the relative positions of these points is given in Table~\ref{tab:tenses}.

While each tensed verb involves a speech, event and reference time, multiple verbs may share one or more of these points. For example, all narrative in a news article usually has the same speech time (that of document creation). Further, two events linked by a temporal conjunction (e.g. \emph{after}) are very likely to share the same reference time.

\small
\begin{table*}
\begin{tabular}{l l l l l}
\hline
        \emph{Relation} & \emph{Reichenbach's Tense Name} & \emph{English Tense Name} & \emph{Example} \\
\hline
        E$<$R$<$S & Anterior past & Past perfect & \emph{I had slept} \\
        E=R$<$S & Simple past & Simple past & \emph{I slept} \\
        R$<$E$<$S & Posterior past &  & \emph{I expected that ..} \\
        R$<$S=E &  &  & \emph{I would sleep} \\
        R$<$S$<$E &  &  &  \\
        E$<$S=R & Anterior present & Present perfect & \emph{I have slept} \\
        S=R=E & Simple present & Simple present & \emph{I sleep} \\
        S=R$<$E & Posterior present & Simple future & \emph{I will sleep (Je vais dormir)} \\
        S$<$E$<$R & Anterior future & Future perfect & \emph{I will have slept} \\
        S=E$<$R &  &  & \\
        E$<$S$<$R &  &  & \\
        S$<$R=E & Simple future & Simple future & \emph{I will sleep (Je dormirai)} \\
        S$<$R$<$E & Posterior future &  & \emph{I shall be going to sleep} \\
\hline
\end{tabular}
\caption[Reichenbach's tenses]{Reichenbach's tenses; from~\newcite{mani2005language}}
\label{tab:tenses}
\end{table*}
\normalsize

From Table~\ref{tab:tenses}, we can see that conventionally English only distinguishes six tenses. Therefore, some English tenses will suggest more than one arrangement of $S$, $E$ and $R$. Reichenbach's tense names suffer from this ambiguity too, but to a much lesser degree. When following Reichenbach's tense names, it is the case that for past tenses, $R$ always occurs before $S$; in the future, $R$ is always after $S$; and in the present, $S$ and $R$ are simultaneous. Further, ``anterior" suggests $E$ before $R$, ``simple" that $R$ and $E$ are simultaneous, and ``posterior" that $E$ is after $R$. The flexibility of this model permits the full set of available tenses~\cite{song1988interpretation}, and this is sufficient to account for the observed tenses in many languages.

Our goal is to define an annotation that can describe $S$, $E$ and $R$ (speech, event and reference time) throughout a discourse. The lexical entities that these times are attached to are verbal events and temporal expressions. Therefore, our annotation needs to locate these entities in discourse, and make the associated time points available.

\subsection{Special properties of the reference point}
The reference point $R$ has two special uses. When sentences or clauses are combined, grammatical rules require tenses to be adjusted. These rules operate in such a way that the reference point is the same in all cases in the sequence. Reichenbach names this principle \textbf{permanence of the reference point}.

Secondly, when temporal expressions (such as a TimeML TIMEX3 of type DATE, but not DURATION) occur in the same clause as a verbal event, the temporal expression does not (as one might expect) specify event time $E$, but instead is used to position reference time $R$. This principle is named \textbf{positional use of the reference point}.

\subsection{Context and the time points}
In the linear order that events and times occur in discourse, speech and reference points persist until changed by a new event or time. That is, the reference time from one sentence will roll over to the next sentence, until it is repositioned explicitly by a tensed verb or time. To cater for subordinate clauses in cases such as reported speech, we add a caveat -- $S$ and $R$ persist as a discourse is read in textual order, \underline{for each context}. We can define a context as an environment in which events occur, such as the main body of the document, reported speech, or the conditional world of an \emph{if} clause~\cite{hornstein1990time}. For example:

\begin{example}
\emph{Emmanuel had said ``This will explode!", but changed his mind.}
\label{exp:srpersist}
\end{example}

Here, \emph{said} and \emph{changed} share speech and reference points. Emmanuel's statement occurs in a separate context, which the opening quote instantiates, ended by the closing quote (unless we continue his reported speech later), and begins with an $S$ that occurs at the same time as \emph{said}'s $E$. This persistence must be explicitly stated in RTMML.

\subsection{Capturing the time points with TimeML}

TimeML is a rich, developed standard for temporal annotation. There exist valuable resources annotated with TimeML that have withstood significant scrutiny. However TimeML does not address the issue of annotating Reichenbach's tense model with the goal of understanding reference time or creating resources that enable detailed examination of the links between verbal events in discourse.

Although TimeML permits the annotation of tense for \texttt{<EVENT>}s, it is not possible to unambiguously map its tenses to Reichenbach's model. This restricts how well we can reason about verbal events using TimeML-annotated documents. Of the usable information for mapping TimeML annotations to Reichenbach's time points, TimeML's \texttt{tense} attribute describes the relation between $S$ and $E$, and its \texttt{aspect} attribute can distinguish between \texttt{PERFECTIVE} and \texttt{NONE} -- that is, between $E < R$ and a conflated class of $(E = R) \vee (R < E)$. Cases where $R < E$ are often awkward in English (as in Table~\ref{tab:tenses}), and may even lack a distinct syntax; the French \emph{Je vais dormir} and \emph{Je dormirai} both have the same TimeML representation and both translate to \emph{I will sleep} in English, despite having different time point arrangements.

It is not possible to describe or build relations to reference points at all in TimeML. It may be possible to derive the information about $S$, $E$ and $R$ directly represented in our scheme from a TimeML annotation, though there are cases -- especially outside of English -- where it is not possible to capture the full nuance of Reichenbach's model using TimeML. An RTMML annotation permits simple reasoning about reference time, and assist the labelling of temporal links between verb events in cases where TimeML's tense and aspect annotation is insufficient. This is why we propose an annotation, and not a technique for deriving $S$, $E$, and $R$ from TimeML.

\section{Overview of RTMML}
\label{rtmml}

The annotation schema RTMML is intended to describe the verbal event structure detailed in~\newcite{reichenbach1947tenses}, in order to permit the relative temporal positioning of reference, event, and speech times. A simple approach is to define a markup that only describes the information that we are interested in, and can be integrated with TimeML. For expositional clarity we use our own tags but it is possible (with minor modifications) to integrate them with TimeML as an extension to the standard.

Our procedure is as follows. Mark all times and verbal events (e.g. TimeML TIMEX3s and those EVENTs whose lexical realisation is a verb) in a discourse, as $T_1 .. T_n$ and $V_1 .. V_n$ respectively. We mark times in order to resolve positional uses of the reference point. For each verbal event $V_i$, we may describe or assign three time points $S_i$, $E_i$, and $R_i$. Further, we will relate $T$, $S$, $E$ and $R$ points using disjunctions of the operators $<$, $=$ and $>$. It is not necessary to define a unique set of these points for each verb -- in fact, linking them across a discourse helps us temporally order events and track reference time. We can also define a ``discourse creation time," and call this $S_{D}$.

\begin{example}
\emph{John said, ``Yes, we have left".}
\label{sentence}
\end{example}
If we let \emph{said} be $V_1$ and \emph{left} be $V_2$:

\begin{itemize*}
\item $S_1 = S_{D}$
\end{itemize*}

From the tense of $V_1$ (simple past), we can say:

\begin{itemize*}
\item $R_1 = S_1$
\item $E_1 < R_1$
\end{itemize*}

As $V_2$ is reported speech, it is true that:

\begin{itemize*}
\item $S_2 = E_1$
\end{itemize*}

Further, as $V_2$ is anterior present:

\begin{itemize*}
\item $R_2 = S_2$
\item $E_2 < R_2$
\end{itemize*}

As the $=$ and $<$ relations are transitive, we can deduce an event ordering $E_2 < E_1$.

\subsection{Annotation schema}

The annotation language we propose is called RTMML, for Reichenbach Tense Model Markup Language. We use standoff annotation. This keeps the text uncluttered, in the spirit of \emph{ISO LAF} and \emph{ISO SemAF-Time}. Annotations reference tokens by index in the text, as can be seen in the examples below. Token indices begin from zero.  We explicitly state the segmentation plan with the \texttt{<seg>} element, as described in~\newcite{lee2010towards} and \emph{ISO DIS 24614-1 WordSeg-1}.

The general speech time of a document is defined with the \texttt{<doc>} element, which has one or two attributes: an ID, and (optionally) \texttt{@time}. The latter may have a normalised value, formatted according to TIMEX3~\cite{boguraev2005timeml} or TIDES~\cite{ferro2005tides}, or simply be the string \texttt{now}. 

Each \texttt{<verb>} element describes a tensed verbal group in a discourse. The \texttt{@target} attribute references token offsets; it has the form \texttt{target="\#token0"} or \texttt{target="\#range(\#token7,\#token10)"} for a 4-token sequence. Comma-separated lists of offsets are valid, for situations where verb groups are non-contiguous. Every \texttt{verb} has a unique value in its \texttt{@id} attribute. The tense of a verb group is described using the attributes \texttt{@view} (with values \emph{simple}, \emph{anterior} or \emph{posterior}) and \texttt{@tense} (\emph{past}, \emph{present} or \emph{future}).

The \texttt{<verb>} element has optional \texttt{@s}, \texttt{@e} and \texttt{@r} attributes; these are used for directly linking a verb's speech, event or reference time to a time point specified elsewhere in the annotation. One can reference document creation time with a value of \texttt{doc} or a temporal expression with its id (for example, \texttt{t1}). To reference the speech, event or reference time of other verbs, we use hash references to the event followed by a dot and then the character \texttt{s}, \texttt{e} or \texttt{r}; e.g., \texttt{v1}'s reference time is referred to as \texttt{\#v1.r}.

As every tensed verb always has exactly one $S$, $E$ and $R$, and these points do not hold specific values or have a position on an absolute scale, we do not attempt to directly annotate them or place them on an absolute scale. One might think that the relations should be expressed in XML links; however this requires reifying time points when the information is stored in the relations between time points, so we focus on the relations between these points for each \texttt{<verb>}. To capture these internal relations (as opposed to relations between the $S$, $E$ and $R$ of different verbs), we use the attributes \texttt{se}, \texttt{er} and \texttt{sr}. These attributes take a value that is a disjunction of $<$, $=$ and $>$.

Time-referring expressions are annotated using the \texttt{<timerefx>} element. This has an \texttt{@id} attribute with a unique value, and a \texttt{@target}, as well as an optional \texttt{@value} which works in the same was as the \texttt{<doc>} element's \texttt{@time} attribute.

\begin{table}
\begin{tabular}{l c}
\hline
Relation name & Interpretation \\
\hline
\textsc{positions}   & $T_a = R_b$ \\
\textsc{same\_timeframe}  & $R_a = R_b [, R_c, .. R_x]$ \\
\textsc{reports}     & $E_a = S_b$ \\
\hline
\end{tabular}
\caption{The meaning of a certain link type between verbs or times a and b.}
\label{tab:linktypes}
\end{table}

\footnotesize
\begin{verbatim}
<rtmml>
Yesterday, John ate well.
 <seg type="token" />
 <doc time="now" />
 <timerefx xml:id="t1" target="
       #token0" />
 <verb xml:id="v1" target="#token3" 
       view="simple" tense="past"
       sr=">" er="=" se=">"
       r="t1" s="doc" />
 </rtmml>
\end{verbatim}
\normalsize

In this example, we have defined a time \emph{Yesterday} as \texttt{t1} and a verbal event \emph{ate} as \texttt{v1}. We have categorised the tense of \texttt{v1} within Reichenbach's nomenclature, using the \texttt{verb} element's \texttt{@view} and \texttt{@tense} attributes.

Next, we directly describe the reference point of \texttt{v1}, as being the same as the time \texttt{t1}. Finally, we say that this verb is uttered at the same time as the whole discourse -- that is, $S_{v1} = S_{D}$. In RTMML, if the speech time of a verb is not otherwise defined (directly or indirectly) then it is $S_{D}$. In cases of multiple voices with distinct speech times, if a speech time is not defined elsewhere, a new one may be instantiated with a string label; we recommend the formatting \emph{s}, \emph{e} or \emph{r} followed by the verb's ID.

This sentence includes a positional use of the reference point, annotated in \texttt{v1} when we say \texttt{r="t1"}. To simplify the annotation task, and to verbosely capture a use of the reference point, RTMML permits an alternative annotation with the \texttt{<rtmlink>} element. This element takes as arguments a relation and a set of times and/or verbs. Possible relation types are \textsc{positions}, \textsc{same\_timeframe} (annotating permanence of the reference point) and \textsc{reports} for reported speech; the meanings of these are given in Table~\ref{tab:linktypes}. In the above markup, we could replace the \texttt{<verb>} element with the following:

\footnotesize
\begin{verbatim}
<verb xml:id="v1" target="#token3" 
      view="simple" tense="past"
      sr=">" er="=" se=">" s="doc" />
<rtmlink xml:id="l1" type="POSITIONS">
  <link source="#t1" />
  <link target="#v1" />
</rtmlink>
\end{verbatim}
\normalsize

When more than two entities are listed as targets, the relation is taken as being between an optional \texttt{source} entity and each of the \texttt{target} entities. Moving inter-verbal links to the \texttt{<rtmlink>} element helps fulfil \emph{TEI p5} and the \emph{LAF} requirements that referencing and content structures are separated. Use of the \texttt{<rtmlink>} element is not compulsory, as not all instances of positional use or permanence of the reference point can be annotated using it; Reichenbach's original account gives an example in German.

\subsection{Reasoning and inference rules}

Our three relations $<$, $=$ and $>$ are all transitive. A minimal annotation is acceptable. The $S$, $E$ and $R$ points of all verbs, $S_{D}$ and all $T$s can represent nodes on a graph, connected by edges labelled with the relation between nodes. 

To position all times in a document with maximal accuracy, that is, to label as many edges in such a graph as possible, one can generate a closure by means of deducing relations. An agenda-based algorithm is suitable for this, such as the one given in~\newcite{setzer2005role}.

\subsection{Integration with TimeML}

To use RTMML as an ISO-TimeML extension, we recommend that instead of annotating and referring to \texttt{<timerefx>}s, one refers to \texttt{<TIMEX3>} elements using their \texttt{tid} attribute; references to \texttt{<doc>} will instead refer to a \texttt{<TIMEX3>} that describes document creation time. The attributes of \texttt{<verb>} elements (except \texttt{xml:id} and \texttt{target}) may be be added to \texttt{<MAKEINSTANCE>} or \texttt{<EVENT>} elements, and \texttt{<rtmlink>}s will refer to event or event instance IDs.

\section{Examples}
\label{examples}

In this section we will give developed examples of the RTMML notation, and show how it can be used to order events and position events on an external temporal scale.

\subsection{Annotation example}
Here we demonstrate RTMML annotation of two short pieces of text.

\subsubsection{Fiction}
From \emph{David Copperfield} by Charles Dickens:
\begin{example}
\emph{When he had put up his things for the night he took out his flute, and blew at it, until I almost thought he would gradually blow his whole being into the large hole at the top, and ooze away at the keys.}
\label{copperfield-plain}
\end{example}

\begin{figure*}
\scriptsize
\begin{multicols}{3}{
\begin{verbatim}
 <doc time="1850" mod="BEFORE" />
 <!-- had put -->
 <verb xml:id="v1"
   target="#range(#token2,#token3)"
   view="anterior" tense="past" />
 <!-- took -->
 <verb xml:id="v2" target="#token11"
   view="simple" tense="past" />
 <!-- blew -->
 <verb xml:id="v3" target="#token17"
   view="simple" tense="past" />
 <!-- thought -->
 <verb xml:id="v4" target="#token24"
   view="simple" tense="past" />
 <!-- would gradually blow -->
 <verb xml:id="v5" 
   target="#range(#token26,#token28)"
   view="posterior" tense="past"
   se="=" er=">" sr=">" 
   r="#v4.e" />
 <!-- ooze -->
 <verb xml:id="v6"
   target="#range(#token26,#token28)"
   view="posterior" tense="past"
   se="=" er=">" sr=">" />
 <rtmlink xml:id="l1" 
   type="SAME_TIMEFRAME">
   <link target="#v1" />
   <link target="#v2" />
   <link target="#v3" />
   <link target="#v4" />
 </rtmlink>
 <rtmlink xml:id="l2"
   type="SAME_TIMEFRAME">
   <link target="#v5" />
   <link target="#v6" />
 </rtmlink>
\end{verbatim}
}
\end{multicols}
\normalsize
\caption{RTMML for a passage from David Copperfield.}
\label{fig:copper}
\end{figure*}

We give RTMML for the first five verbal events from Example~\ref{copperfield-plain} RTMML in Figure~\ref{fig:copper}. The fifth, \texttt{v5}, exists in a context that is instantiated by \texttt{v4}; its reference time is defined as such. We can use one \texttt{link} element to show that \texttt{v2}, \texttt{v3} and \texttt{v4} all use the same reference time as \texttt{v1}. The temporal relation between event times of \texttt{v1} and \texttt{v2} can be inferred from their shared reference time and their tenses; that is, given that \texttt{v1} is anterior past and \texttt{v2} simple past, we know $E_{v1} < R_{v1}$ and $E_{v2} = R_{v2}$. As our \texttt{<rtmlink>} states $R_{v1} = R_{v2}$, then $E_{v1} < E_{v2}$. Finally, \texttt{v5} and \texttt{v6} happen in the same context, described with a second \textsc{same\_timeframe} link.

\subsubsection{Editorial news}
From an editorial piece in TimeBank~\cite{pustejovsky2003timebank} (AP900815-0044.tml):
\begin{example}
\emph{Saddam appeared to accept a border demarcation treaty he had rejected in peace talks following the August 1988 cease-fire of the eight-year war with Iran.}
\label{timeml-plain}
\end{example}
\footnotesize
\begin{verbatim}
 <doc time="1990-08-15T00:44" />
 <!-- appeared -->
 <verb xml:id="v1" target="#token1"
   view="simple" tense="past" />
 <!-- had rejected -->
 <verb xml:id="v2"
   target="#range(#token9,#token10)"
   view="anterior" tense="past" />
 <rtmlink xml:id="l1" 
   type="SAME_TIMEFRAME">
   <link target="#v1" />
   <link target="#v2" />
 </rtmlink>
\end{verbatim}
\normalsize

Here, we relate the simple past verb \emph{appeared} with the anterior past (past perfect) verb \emph{had rejected}, permitting the inference that the first verb occurs temporally after the second. The corresponding TimeML (edited for conciseness) is:

\footnotesize
\begin{verbatim}
Saddam <EVENT eid="e74" class="I_STATE">
appeared</EVENT> to accept a border 
demarcation treaty he had <EVENT eid="e77"
class="OCCURRENCE">rejected</EVENT>

<MAKEINSTANCE eventID="e74" eiid="ei1568"
 tense="PAST" aspect="NONE" polarity="POS"
 pos="VERB"/>
<MAKEINSTANCE eventID="e77" eiid="ei1571"
 tense="PAST" aspect="PERFECTIVE"
 polarity="POS" pos="VERB"/>
\end{verbatim}
\normalsize

In this example, we can see that the TimeML annotation includes the same information, but a significant amount of other annotation detail is present, cluttering the information we are trying to see. Further, these two \texttt{<EVENT>} elements are not directly linked, requiring transitive closure of the network described in a later set of \texttt{<TLINK>} elements, which are omitted here for brevity.

\subsection{Linking events to calendrical references}

RTMML makes it possible to precisely describe the nature of links between verbal events and times, via positional use of the reference point. We will link an event to a temporal expression, and suggest a calendrical reference for that expression, allowing the events to be placed on a calendar. Consider the below text, from wsj\_0533.tml in TimeBank.

\begin{example}
\emph{At the close of business Thursday, 5,745,188 shares of Connaught and C\$44.3 million face amount of debentures, convertible into 1,826,596 common shares, had been tendered to its offer.}
\label{exp:positiontimebank}
\end{example}

\footnotesize
\begin{verbatim}
 <doc time="1989-10-30" />
 <!-- close of business Thursday -->
 <timerefx xml:id="t1"
   target="#range(#token2,#token5)" />
 <!-- had been tendered -->
 <verb xml:id="v1"
   target="#range(#token25,#token27)"
   view="anterior" tense="past" />
 <rtmlink xml:id="l1" target="#t1 #v1">
   <link target="#t1" />
   <link target="#v1" />
 </rtmlink>
\end{verbatim}
\normalsize

This shows that the reference time of \texttt{v1} is \texttt{t1}. As \texttt{v1} is anterior, we know that the event mentioned occurred before \emph{close of business Thursday}. Normalisation is not a task that RTMML addresses, but there are existing methods for deciding which Thursday is being referenced given the document creation date~\cite{mazur2008s}; a time of day for \emph{close of business} may be found in a gazetteer.

\subsection{Comments on annotation}
As can be seen in Table~\ref{tab:tenses}, there is not a one-to-one mapping from English tenses to the nine specified by Reichenbach. In some annotation cases, it is possible to see how to resolve such ambiguities. Even if view and tense are not clearly determinable, it is possible to define relations between $S$, $E$ and $R$; for example, for arrangements corresponding to the simple future, $S < E$. In cases where ambiguities cannot be resolved, one may annotate a disjunction of relation types; in this example, we might say ``$S < R$ or $S = R$" with \texttt{sr="<="}.

Contexts seem to have a shared speech time, and the $S-R$ relationship seems to be the same throughout a context. Sentences which contravene this (e.g. \emph{``By the time I ran, John will have arrived"}) are rather awkward.

RTMML annotation is not bound to a particular language. As long as a segmentation scheme (e.g. WordSeg-1) is agreed and there is a compatible system of tense and aspect, the model can be applied and an annotation created.

\section{Conclusion and Future Development}
\label{conclusion}
Being able to recognise and represent reference time in discourse can help in disambiguating temporal reference, determining temporal relations between events and in generating appropriately tensed utterances. A first step in creating computational tools to do this is to develop an annotation schema for recording the relevant temporal information in discourse. To this end we have presented RTMML, our annotation for Reichenbach's model of tense in natural language. 

We do not intend to compete with existing languages that are well-equipped to annotate temporal information in documents; RTMML may be integrated with TimeML. What is novel in RTMML is the ability to capture the abstract parts of tense in language. We can now annotate Reichenbach's time points in a document and then process them, for example, to observe interactions between temporal expressions and events, or to track reference time through discourse. This is not directly possible with existing annotation languages.

There are some extensions to Reichenbach's model of the tenses of verbs, which RTMML does not yet cater for. These include the introduction of a reference interval, as opposed to a reference point, from~\newcite{dowty1979word}, and Comrie's suggestion of a second reference point in some circumstances~\cite{comrie1985tense}. RTMML should cater for these extensions.

Further, we have preliminary annotation tools and have begun to create a corpus of annotated texts that are also in TimeML corpora. This will allow a direct evaluation of how well TimeML can represent Reichenbach's time points and their relations. To make use of Reichenbach's model in automatic annotation, given a corpus, we would like to apply machine learning techniques to the RTMML annotation task. Work in this direction should enable us to label temporal links and to anchor time expressions with complete accuracy where other systems have not succeeded.

\section{Acknowledgements}
The authors would like to thank David Elson for his valuable comments. The first author would also like to acknowledge the UK Engineering and Physical Science Research Council's support in the form of a doctoral studentship.

\nocite{iso2009wordseg}
\nocite{iso2009semaftime}
\nocite{iso2009laf}


\end{document}